\documentclass[journal]{IEEEtran}
\ifCLASSINFOpdf
\else
\fi
\usepackage{times}
\usepackage{float}
\usepackage{booktabs}
\usepackage{amsmath}
\usepackage{amssymb}
\usepackage{graphicx}
\usepackage{subfigure}
\usepackage{amsfonts}
\usepackage{amsmath}
\usepackage{multirow}
\usepackage{xcolor}

\usepackage{amsmath}
\usepackage{amssymb}
\usepackage{amsthm}
\usepackage{cases}
\usepackage{bm}
\usepackage{algorithm}
\usepackage{algorithmic}
\usepackage{cases}
\usepackage{subfigure}
\usepackage{multirow}
\usepackage{xcolor} 
\usepackage{enumerate}
\usepackage{etoolbox}
\usepackage{boxedminipage} 
\usepackage{booktabs} 
\usepackage{graphicx}

\newcommand{\MyMapTemplatePrefix}[4]{\expandafter#1\csname#3#4\endcsname{#2{#4}}}
\newcommand{\MyMapTemplatePrefixNew}[5]{\expandafter#1\csname#4#5\endcsname{#2{#3{#5}}}}
\forcsvlist{\MyMapTemplatePrefix {\def} {\mathbf} {}} {A,B,C,D,E,F,G,H,I,J,K,L,M,N,O,P,Q,R,S,T,U,V,W,X,Y,Z} 
\forcsvlist{\MyMapTemplatePrefix {\def} {\mathbf} {}} {a,b,c,d,e,f,g,h,i,j,k,l,m,n,o,p,q,r,s,t,u,v,w,x,y,z,1,0}
\forcsvlist{\MyMapTemplatePrefix {\def} {\widetilde} {wt}} {A,B,C,D,E,F,G,H,I,J,K,L,M,N,O,P,Q,R,S,T,U,V,W,X,Y,Z}
\forcsvlist{\MyMapTemplatePrefix {\def} {\widetilde} {wt}} {a,b,c,d,e,f,g,h,i,j,k,l,m,n,o,p,q,r,s,t,u,v,w,x,y,z} 
\forcsvlist{\MyMapTemplatePrefixNew {\def} {\widetilde}{\mathbf} {tb}} {A,B,C,D,E,F,G,H,I,J,K,L,M,N,O,P,Q,R,S,T,U,V,W,X,Y,Z}
\forcsvlist{\MyMapTemplatePrefixNew {\def} {\widetilde}{\mathbf} {tb}} {a,b,c,d,e,f,g,h,i,j,k,l,m,n,o,p,q,r,s,t,u,v,w,x,y,z}
\forcsvlist{\MyMapTemplatePrefix {\def} {\widehat} {wh}} {A,B,C,D,E,F,G,H,I,J,K,L,M,N,O,P,Q,R,S,T,U,V,W,X,Y,Z}
\forcsvlist{\MyMapTemplatePrefix {\def} {\widehat} {wh}} {a,b,c,d,e,f,g,h,i,j,k,l,m,n,o,p,q,r,s,t,u,v,w,x,y,z}
\forcsvlist{\MyMapTemplatePrefixNew {\def} {\widehat}{\mathbf} {hb}} {A,B,C,D,E,F,G,H,I,J,K,L,M,N,O,P,Q,R,S,T,U,V,W,X,Y,Z}
\forcsvlist{\MyMapTemplatePrefixNew {\def} {\widehat}{\mathbf} {hb}} {a,b,c,d,e,f,g,h,i,j,k,l,m,n,o,p,q,r,s,t,u,v,w,x,y,z}
\forcsvlist{\MyMapTemplatePrefix {\def} {\mathcal}{mc}} {A,B,C,D,E,F,G,H,I,J,K,L,M,N,O,P,Q,R,S,T,U,V,W,X,Y,Z}
\forcsvlist{\MyMapTemplatePrefix {\def} {\mathbb} {mb}} {A,B,C,D,E,F,G,H,I,J,K,L,M,N,O,P,Q,R,S,T,U,V,W,X,Y,Z}
\forcsvlist{\MyMapTemplatePrefix {\DeclareMathOperator} {} {} } {tr,diag,sgn}
\def\tp{^\top}
\def\st{\text{s.t.~}}

\usepackage[pagebackref=true,breaklinks=true,letterpaper=true,colorlinks,linkcolor = black,citecolor = black,citecolor=black,bookmarks=false]{hyperref}
\newcommand{\changeurlcolor}[1]{\hypersetup{urlcolor=#1}} 
\begin{document}
%
\title{Localizing Anomalies from Weakly-Labeled Videos}

%
%
%
\author{Hui Lv, Chuanwei Zhou, Zhen Cui\thanks{Corresponding author: Zhen Cui, zhen.cui@njust.edu.cn.}, Chunyan Xu, Yong Li, Jian Yang
\thanks{Email address: (hubrthui, cwzhou, cyx, yong.li, csjyang) @njust.edu.cn (H. Lv, C. Zhou, C. Xu, Y. Li, J. Yang).}
\thanks{H. Lv, C. Zhou, Z. Cui, C. Xu, Y. Li and J. Yang are from School of Computer Science and Engineering, Nanjing University of Science and Technology, Nanjing, Jiangsu, China.}
}

\maketitle

\begin{abstract}
Video anomaly detection under video-level labels is currently a challenging task. 
Previous works have made progresses on discriminating whether a video sequence contains anomalies. However, most of them fail to accurately localize the anomalous events within videos in the temporal domain. 
In this paper, we propose a Weakly Supervised Anomaly Localization (WSAL) method focusing on temporally localizing anomalous segments within anomalous videos.
Inspired by the appearance difference in anomalous videos, the evolution of adjacent temporal segments is evaluated for the localization of anomalous segments.
To this end, a high-order context encoding model is proposed to not only extract semantic representations but also measure the dynamic variations so that the temporal context could be effectively utilized.
In addition, in order to fully utilize the spatial context information, the immediate semantics are directly derived from the segment representations. The dynamic variations as well as the immediate semantics, are efficiently aggregated to obtain the final anomaly scores.
An enhancement strategy is further proposed to deal with noise interference and the absence of localization guidance in anomaly detection.
Moreover, to facilitate the diversity requirement for anomaly detection benchmarks, we also collect a new 
traffic anomaly (TAD) dataset which specifies in the traffic conditions, differing greatly from the current popular anomaly detection evaluation benchmarks.\footnote{The dataset and the benchmark test codes, as well as experimental results, are made public on \changeurlcolor{black}{\url{http://vgg-ai.cn/pages/Resource/}} and \changeurlcolor{black}{\url{https://github.com/ktr-hubrt/WSAL}}.}
Extensive experiments are conducted to verify the effectiveness of different components, and our proposed method achieves new state-of-the-art performance on the UCF-Crime and TAD datasets.
\end{abstract}


%
\IEEEpeerreviewmaketitle
\section{Introduction}
\label{section:intro}

Anomaly detection, which aims to recognize those behaviors or appearance patterns that do not conform to usual patterns~\cite{chandola2009anomaly,adam2008robust,benezeth2009abnormal}, is of great importance for the alarm of potential risks or dangers. With the large-scale deployment of surveillance, an urgent requirement of intelligent systems is to automatically filter out possibly abnormal events.

\begin{figure}[!t] 
	\centering 
	\includegraphics[width=0.45\textwidth]{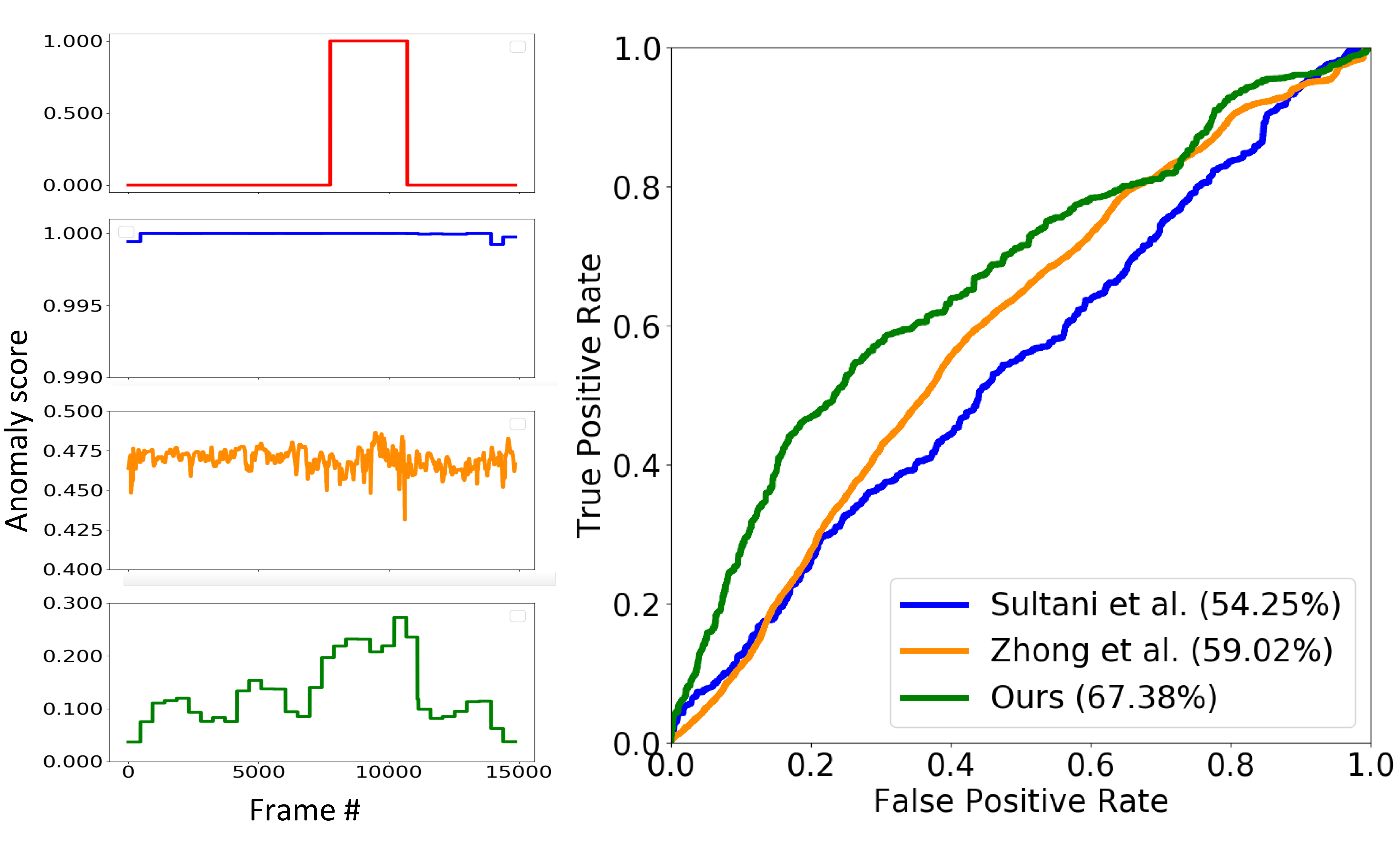}
	\caption{Anomaly localization comparisons. \textbf{Left:} A comparison of \textit{Burglary} case on UCF-Crime (x-axis corresponds to frames and y-axis corresponds to the anomaly score.). Groundtruth is shown in the top-left, following by three  methods: Sultani et al.~\cite{sultani2018real}, Zhong et al.~\cite{zhong2019graph} and ours. \textbf{Right:}  ROC curves of frame-level anomaly localization on all anomaly videos}
	\label{Fig.rc}
	\vspace{-0.3cm}
\end{figure}
Anomaly detection is typically tackled under constrained supervision that only normal data or limited annotations are provided in the training phase.~\cite{zhong2019graph,zhu2019motion,sultani2018real,liu2018future,luo2017revisit,chong2017abnormal}. 
As anomalous events rarely happen in real-life situations, which brings in the scarcity of annotations, several methods~\cite{liu2018future,luo2017revisit,chong2017abnormal} have been proposed to model the shared pattern among normal videos in the training phase and detect the outliers as anomalies during testing. However, these methods often fail in identifying anomalies when facing complicated or unseen scenes. Recently, researchers~\cite{sultani2018real} select to leverage the video-level labels for developing robust anomaly detectors. The release of UCF-Crime dataset \cite{sultani2018real} activates this direction which encourages the detectors to take the best of the weak signals of video-level. Although a large gain has been observed in this domain~\cite{zhu2019motion,zhong2019graph}, it still lacks an efficient way to temporally localize anomalous frames.

In previous methods, the performances on the overall test set are calculated and reported as the evaluation results. However, in this case the temporal anomaly localization capability of detectors is somewhat unrevealed. Since the whole test set contains both the normal and anomaly videos, the superior performance on normal videos conceals the poor accuracy of anomaly localization within anomalous videos. To reveal the problem therein, we conduct statistic analysis on the anomaly data of UCF-Crime test set. ROC curves of two state-of-the-art (SOTA) methods, as well as ours, are plotted in \color{black}{Figure~\ref{Fig.rc}}. The details of corresponding metrics can be found in Section~\ref{EM}. A test sample (video name: \textit{Burglary079}) is also shown in the left part of the figure. We can find that the localization accuracy of the two methods on anomalous videos are 54.25\% and 59.02\% respectively, in term of AUC. It is worth mentioning that an AUC of 50\% can be obtained by random binary prediction of anomalies.
To sum up, there exists a large space for improving the temporal localization of anomalies.

To facilitate the localization property of anomaly detection, we propose a Weak-Supervised Anomaly Localization (WSAL) method to detect anomalies with video-level labels. In our WSAL model, we investigate into two aspects of the anomaly, which are the semantic and context. The anomalies are defined as the uncommon activities that differ from the usual pattern. Thus, the extracted semantics can act as a direct cue to infer anomalies. Based on this point,
existing methods \cite{sultani2018real,zhong2019graph} treat each video as frame-by-frame images or direct optical flows and extract fine-grained semantic representations for further anomaly detection. 
While in this manner, the temporal evolution across consecutive frames is not adequately exploited.
For example, in the long temporal domain, a sudden change of the dynamic variation uncovers the anomaly itself.

On the other hand, owing to rough supervisory signals of video level, anomaly detectors are prone to false alarms or missed detection. For instances, the drastic environment changes as well as noise interruptions caused by hardware failure may lead to unwanted high probabilities from anomaly detectors. These influences in long and untrimmed videos ought to be suppressed or excluded from the anomalies. Toward this end, we put forward a noise stimulation strategy to tackle inevitable interference lying in untrimmed videos, whose quality can not be guaranteed. Moreover, we introduce hand-crafted anomalies, similar to actual anomalies, to provide pseudo location signals as guidance for the model learning process.
Above two strategies make up for our enhancement strategy to boost the weakly-supervised learning and strengthen the robustness of anomaly detection.
Thoroughly, we equip raw video data with the augments of video noises and hand-crafted anomalies.
As a consequence, the weak labels are expanded with pseudo location signals as auxiliary.

So far, there are few datasets available for anomaly detection, most of them are with small-scale or constrained scenarios, like UCSD Peds \cite{li2013anomaly}, Avenue \cite{lu2013abnormal}, ShanghaiTech \cite{luo2017revisit}, and Street Scene \cite{ramachandra2020street}.
Also, these datasets are initially used for semi-supervised anomaly detection with normal training samples.
For the problem under video-level scenario, only UCF-Crime \cite{sultani2018real} dataset is now available publicly to our knowledge.
Thus, we build a new large-scale traffic anomaly detection (TAD) dataset with long surveillance videos under traffic scene.
The proposed dataset consists of realistic anomalies on roads with various appearance and motion pattern, which facilitates the diversity requirement for anomaly detection benchmarks. In addition, we implement and compare different SOTA anomaly detection approaches on the UCF-Crime and our TAD dataset. 
We hope the newly collected benchmark will boost the development of anomaly detection in research domain and real-life application.
The main contributions of this paper are as follows:
\begin{enumerate}
  \item Deeply delving into anomaly detection, we propose a weak-labeled anomaly localization method, in which we employ a high-order context encoding model to encode temporal variations as well as high-level semantic information for weak-supervised anomaly detection;
  \item We introduce a weak-supervision enhancement strategy by stimulating video noises and building virtual indicative locations to suppress or exclude those interruption of false-anomaly signals;
  \item We build a new weak-labeled traffic anomaly detection dataset with extensive benchmark tests,  and report the new SOTA results on the proposed TADdataset as well as the UCF-Crime dataset.
\end{enumerate}

The rest parts of the paper are organized as follows: In Section \ref{sec:relatedwork}, we review the literature of anomaly detection in surveillance videos. In Section \ref{sec:TPM} we introduce the proposed WSAL method in details. In Section \ref{sec:exp}, we conduct experiments to compare our proposed method with other SOTA methods as well as elaborated ablation studies to fully analyze different components.
Finally, we conclude the paper in Section \ref{sec:con}.

\section{Related work}
\label{sec:relatedwork}
The techniques of anomaly detection in surveillance videos have long been developed as a tool for mining unusual patterns in videos \cite{zhao2011online,kratz2009anomaly,wu2010chaotic,li2013anomaly,antic2011video}.
The family can be divided into two categories, based on how and how much supervision is accessible. 
The details are discussed in the following. 

Video anomaly detectors are originally designed in an unsupervised manner~\cite{chong2017abnormal,hasan2016learning,morais2019learning,del2016discriminative,park2020learning} that only normal samples are available in the training phase without any labels. They first involve modeling normal behavior and then detecting samples that deviate from it.
Motion trajectory, as one of common basic factors, has been utilized to detect anomalies in \cite{bensch2017spatiotemporal,basharat2008learning,wu2010chaotic}.
Although such methods can be easily implemented and have a fast execution speed, tracking is prone to failure in crowded or cluttered scenes.
An alternative approach is to tackle the original task as a problem of novelty detection, e.g., sparse coding \cite{cong2011sparse,lu2013abnormal,zhao2011online}, distance-based methods \cite{saligrama2012video}, the mixture of dynamic models on texture \cite{mahadevan2010anomaly} and the mixture of probabilistic PCA \cite{kim2009observe}.
These models are generally built on the low-level features (e.g., a histogram of oriented gradients (HOG) and the histogram of oriented flows (HOF)) extracted from densely sampled image patches.
There are also works that improve the tradition approach into VAD, such as in~\cite{yuan2016anomaly}, the authors proposes a spatial localization constrained sparse coding approach for anomaly detection in traffic scenes, which fuses these two aspects of motion orientation and magnitude to obtain a robust detection result. Several recent approaches have investigated the learning-based features using autoencoders \cite{sabokrou2015real,xu2015learning}, which minimize reconstruction errors on the normal patterns in the training process.
Shi et al. \cite{xingjian2015convolutional} have proposed to modify original LSTM with ConvLSTM and used it for precipitation forecasting.
Liu et al. \cite{liu2018future} have designed a future prediction network to infer the coming frames and detect anomalies according to the quality of predicted frames.
Despite the advances in developing unsupervised anomaly detection approaches, these detectors are easily to fall down when dealing with complicated or unseen environments.

Recently, great advance has been witnessed in weak supervision, for example in \cite{wang2019weakly}, the authors introduce weak supervision into adversarial domain adaptation for improving the segmentation performance from synthetic data to real scenes. 
Inspired by them, various methods based on weak supervision situation have been introduced, they employ both normal and abnormal data along with video-level annotations for building robust anomaly detection model \cite{adhiya2009tracking,sultani2018real,he2018anomaly,zhu2019motion,zhong2019graph}.
Among them, Multiple Instance Learning (MIL) is introduced for pattern modeling under weak supervision \cite{sultani2018real,he2018anomaly,zhu2019motion}.
Sultani et al. \cite{sultani2018real} consider anomaly detection as a MIL problem with a novel ranking loss function. Later, by extending it, Zhu et al. \cite{zhu2019motion} introduce the attention mechanism for better localizing anomalies. Due to the absence of anomaly positions in training phase, these two methods cannot predict anomaly frames well. For this, Zhong et al. \cite{zhong2019graph} attempt to construct supervised signals of anomaly positions through iteratively refining them. 
However these methods focus on predicting segment labels while neglecting modeling hidden temporal context information.
Temporal or context aggregation technology has been widely adopted as in~\cite{li2020spatiotemporal} the authors adopt a volumetric structure to effectively synthesize spatiotemporal information of the same target from the current time and history frames to enhance detection. In~\cite{zhang2017robust}, the authors propose two nuclear- and L2,1-norm regularized neighborhood preserving projection methods for extracting representative 2D image features.
While in our work, we not only propose a high-order context encoding structure for temporal context aggregation but also modeling variations through the video sequences as a dynamic cue and incorporate it with semantic cues to better localize anomalies. Besides, we introduce a weak-supervision enhancement strategy to suppress false-anomaly signals.

\section{The Proposed Method}
\label{sec:TPM}
In this section, we will introduce our Weak-Supervised Anomaly Localization (WSAL) method in details. We first give the basic formulation for the anomaly localization and core modules of our WSAL are elaborated thoroughly then.

\subsection{Formulation}
The purpose of anomaly detection is to estimate the anomaly status of a video and localize the anomalies in the video sequence if exist.  
In the weakly supervised scenario, a video sequence $\mcX$ and its corresponding video-level annotation y $\in \{0,1\}$ are given, where the case `y=1' means there exists anomaly in this sequence otherwise `y=0' indicates that there is no anomaly in $\mcX$.
We start with dividing the entire video into several segments with equal lengths, denoted as $\mcX=(\mcX_1, \mcX_2,\cdots, \mcX_m)$. 
The goal of video segmenting is to alleviate computation burden resulting from almost-repetitive video frames. 
For the $i$-th segment $\mcX_i$, we first use a classical convolutional network to extract features for each frame, and the segment feature $x_i$ is obtained by aggregating the features of all frames within the segment. As a consequence, the sequence $\mcX$ could now be represented by the $m$-tuple features $(\x_1, \x_2, \cdots, \x_m).$ We can now use this $m$-tuple to determine whether the current video contains any anomaly or not, in the manner that assigning each segment in the video with an anomaly score, indicating the probability of being anomalous.

To predict the state (normal or abnormal) of a video, we derive a novel function to describe the video by estimating the anomalous margin among a video, formally,
\begin{align}
\label{f1}
\mcS(\mcX) = \max \limits_{i,j=1,\dots, m}f(\psi(\x_{i-k},\dots,\x_{i},\dots,\x_{i+k}),\nonumber\\
\psi(\x_{j-k},\dots,\x_{j},\dots,\x_{j+k})),
\end{align}
where
\begin{itemize}
	\item $\psi$ is a high-order function that encodes an anchored segment as well as its adjacent $2k$ segments in the temporal context. To mine the anomalies, we consider two aspects of information: spatial semantics and temporal variations. The function $\psi$ is modeled with a high-order dynamic regression to generate semantic features and predict variations within local window $[-k,k]$. Please see Section~\ref{sec:dual} for more details.
	\item $f$ is a margin distance metric measuring the anomaly score margin between the segment position $i$ and $j$. The more close the predicted anomaly scores are, the smaller the distance is. 
	\item $\mcS(\cdot)$ is the score of a video that computes the maximum relative distance of pairwise positions. The scores of normal videos are expected to be smaller than anomalous videos. Thus, the maximum-distance strategy constrains entire normal videos more smooth than anomaly videos, which complies with the conventional assumption.  
	\item $\max$ function is chosen to capture the largest score margin, which can represent the extent of abnormalities in a video. Since anomaly scores are all close to zero in a normal video, leading to the score margin with a small value. While in an anomalous video, the anomalies, lying in normal background, will lead to a large score margin.
\end{itemize}

Given a batch of training data $\{\mcX^{(1)},\mcX^{(2)},\cdots,\mcX^{(n)}\}$ and the corresponding video labels $\{y^{(1)},y^{(2)},\cdots,y^{(n)}\}$, we define a margin loss function as:

\begin{align}
\zeta_1&(\{\mcX^{(i)}\}|_{i=1}^n) =\max~\{0,~1-\frac{1}{n_1}\sum_{i=1}^n [\mcS(\mcX^{(i)})|{y^{(i)}=1}]\nonumber\\
&\qquad\qquad\qquad\qquad+\frac{1}{n_0}\sum_{j=1}^n [\mcS(\mcX^{(j)})|{y^{(j)}=0}]~\},\label{eqn:L1}
\end{align}
here $n_1$, $n_0$ are the total amounts of anomaly and normal samples. As the function only depends on video-level labels, the learning process belongs to the case of weak supervision.

In addition, we augment training samples to generate two types of data: noise data $\{\ddot\mcX^{(i)}\}|_{i=1}^{\ddot n}$ and pseudo-location data $\{\breve\mcX^{(i)}\}|_{i=1}^{\breve n}$, where $\ddot n$ and $\breve n$ are the amounts of pseudo samples. The former could help the detector reduce mis-judgement where some noised normal videos are predicted as anomaly labels, whilst the latter provides direct guidance to localize anomalous frames. Let $\{\mcX'\}|_{i=1}^{n'}=\{\ddot\mcX^{(i)}\}|_{i=1}^{\ddot n}\cup\{\breve\mcX^{(i)}\}|_{i=1}^{\breve n}$ denote all augmentation samples, where $n'=\ddot n + \breve n$. Finally, we derive the objective function to optimize as:
\begin{align}
\label{final_formula}
\zeta = \zeta^{\mcO}(\{\mcX^{(i)}\}|_{i=1}^n)+\lambda\zeta^{\mcA}(\{\mcX'^{(i)}\}|_{i=1}^{n'}),
\end{align}
where $\lambda$ is the balance factor between the original and the augmented data. Loss function $\zeta^{\mcO}$, defined on the original weak-labeled data, uses the margin loss $\zeta_1$ in Equation~(\ref{eqn:L1}), and the details will be listed in Section~\ref{sec:dual}. Loss $\zeta^{\mcA}$ is imposed on noise data as well as pseudo-location data, which will be introduced in Section~\ref{sec:aug}.

In testing process, given a video, we obtain the anomaly status of each segment by aggregating the consensus of spatial semantics and dynamic variations defined in the following.

\subsection{High-order Context Encoding}
\label{sec:dual}

Previous approaches \cite{sultani2018real,zhu2019motion,zhong2019graph} directly infer the anomaly scores from input visual features in an intuitive way, while neglecting the guidance of the temporal context for anomaly localization.
Intuitively, the rarely occurred anomalies among the normal patterns will lead to significant changes in the time domain. Therefore, the dynamic variations in the time series are able to indicate the existence of anomalies.
Inspired by this, we propose to leverage the temporal context information for the immediate spatial semantics and dynamic temporal variations, and aggregate both cues for accurately locating anomalies.

In the beginning, we design a High-order Context Encoding (HCE) model to extract high-level semantic features and encode the variations in time series. The input is the feature vectors $({\x_1},\cdots,{\x_m})$ extracted from consecutive segments. The regression process is formulated as:
\begin{equation}
\tbx_{t} = \sum_{j=-k,\cdots, k,~j\neq 0} \W_j\x_{t+j}+ \W_0\x_{t}+\mathbf{b} ,
\end{equation}
where $\W_j$ is a projection function on the $j$-th segment, $\mathbf{b}$ is a bias term. The output encodes the context information of the anchored segment and adjacent segments, i.e., $({\tbx_{t-k}},\cdots,\tbx_{t-1},\x_t,\tbx_{t+1},\cdots,{\tbx_{t+k}})$. 
The intuition is that $t$-th high-order feature vector collects the fruitful information from its $2k$ neighbors, which can facilitate both the mining of immediate spatial semantics and local dynamic variations. Actually the regression can be stacked as a hierarchical structure by taking the output $\tbx$ as the input in a recursive manner. In practice, we find the simple one-layer regression can perform well.

The neighbor size $k$ controls the temporal context modeled in each local segment $\tbx_{t}$. 
Then to exploit the immediate semantic information of the anchored segment, we use a fully connected layer, activated by a Sigmoid function, to obtain an anomaly score. Formally:

\begin{align}
\psi^{sem}(\tbx_t) &= \sigma(\w_{sem}\tbx_t +{b}_{sem}),
\end{align}
where $\psi^{sem}(\tbx_t)$ represents the semantics score, $w_{sem}$ and $b_{sem}$ are the weight and bias of the fully connected layer and $\sigma$ stands for the sigmoid function.

To measure the variation between two adjacent segments, we take the cosine similarity measurement: $\cos(\tbx_{t-1},\tbx_{t})=\tbx_{t-1}\tp\tbx_{t}/(\|\tbx_{t-1}\|^2\|\tbx_{t}\|^2)$. The corresponding distance metric is $1-\cos(\tbx_{t-1},\tbx_{t})$, which has a large value for dramatic variations. Then the second-order discrepancy of local variations is computed as an indicator of anomaly, which becomes:
\begin{align}
\psi^{var}(\tbx_t) = (2-\cos(\tbx_{t-1},\tbx_{t})-\cos(\tbx_{t},\tbx_{t+1})))/4,
\end{align}
where we make the score value divided by four to normalize the scalar into $[0,1]$. 

Then, we obtain the singularity of a sequence from the dual context cues, with the margin measurement $f$ as L1-distance:
\begin{align}
\mcS^{sem}(\mcX) &= \max_{i,j=1,\cdots, m} |\psi^{sem}(\tbx_i)-\psi^{sem}(\tbx_j)|,\\
\mcS^{var}(\mcX) &= \max_{i,j=1,\cdots, m} |\psi^{var}(\tbx_i)-\psi^{var}(\tbx_j)|.
\end{align}

By plugging above singularity tuple into Equation~(\ref{eqn:L1}), the acquired margin losses of the dual context are denoted as $\zeta_1^{sem}$ and $\zeta_1^{var}$.
Since the scores of normal events are targeted to $0$, and those of anomalous are sparse (scarce of anomalies), we place a sparsity constraint on the loss function.
Added with the sparsity constraint of weight $\beta$, the margin loss of dual context becomes:
\begin{align}
\label{eqn:zeta_o}
\zeta^{\mcO} &= \zeta_1^{sem}(\{\mcX^{(i)}\}|_{i=1}^n)+\zeta_1^{var}(\{\mcX^{(i)}\}|_{i=1}^n) \nonumber \\
&+ \frac{\beta}{n} \sum_{i=1}^n\sum_{t=1}^m (|s_t^{sem}|+|s_t^{var}|).
\end{align}

\vspace{-0.3cm}
\begin{figure}[t]
	\centering 
	\includegraphics[width=0.5\textwidth]{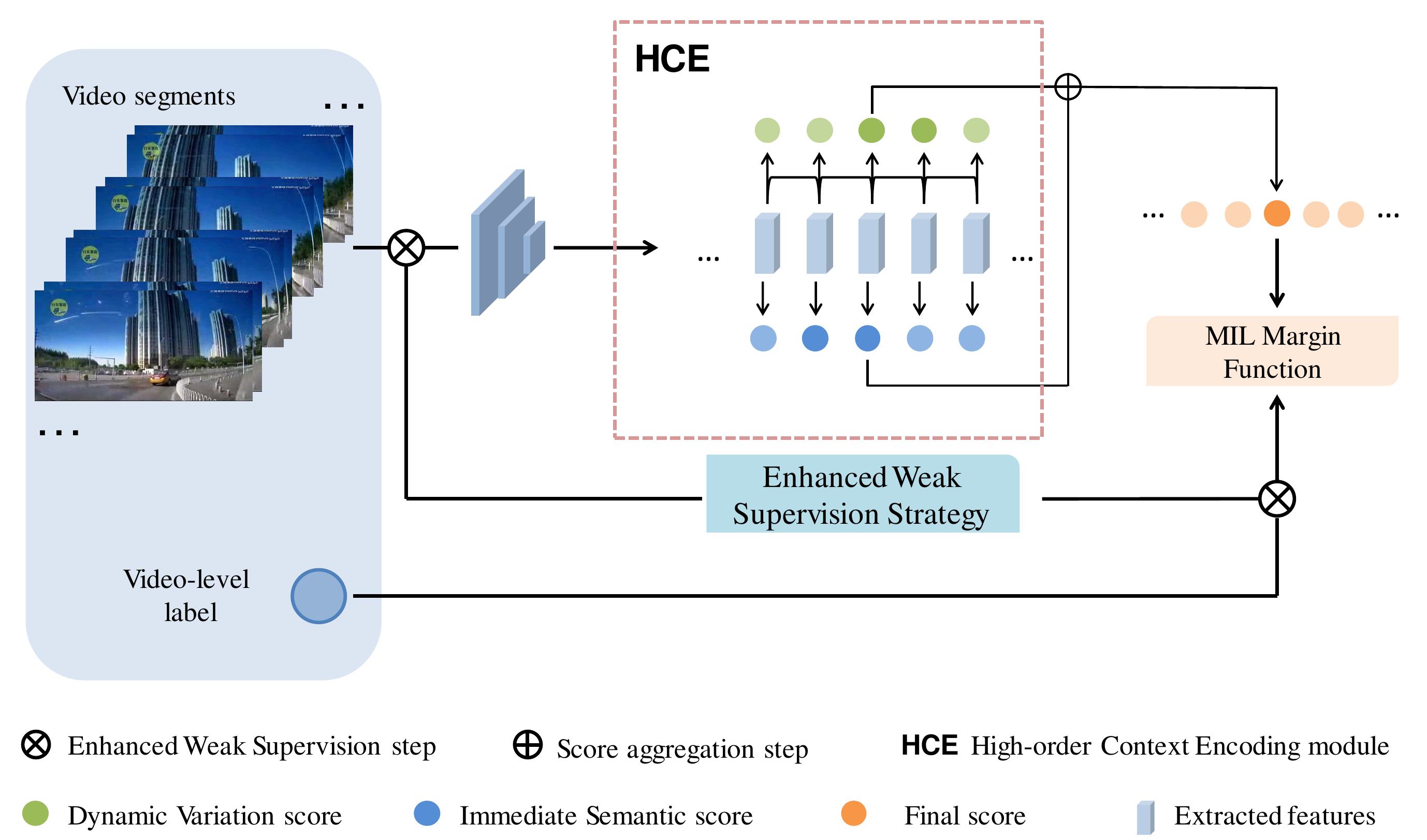}
	\caption{Framework of WSAL. Video clips are organized in segment level and inputted into the backbone model. These extracted features are processed in HCE module to generate anomaly scores from the cues of immediate semantics and dynamic variations. Then the predicted scores are aggregated and supervised in a novel MIL Margin objective function using the video-level labels. In weak supervision, only video-level annotations are available, which lacks accurate temporal location guidance. Motivated by this, we introduce Enhanced Weak Supervision strategy for data augmentation and generating pseudo anomaly signals. Better viewed in color.}
	\label{pip}
\end{figure} 
\subsection{Enhanced Weak Supervision}\label{sec:aug}

\textbf{Noise Simulation}. As is mentioned in Section \ref{section:intro}, noises in videos lead to serious interference for anomaly detection, especially localization.
Due to the unavoidable external factors, it tends to exist noisy artifacts such as lens jitter in the videos which is going to result in misjudgments.
To mitigate this issue, we introduce a noise simulation strategy in which we fuse the raw videos with varying degrees of video noises, such as blur, picture interruption as well as lens jitter.
Specifically, we augment the normal video sequences with three kinds of video noise simulations, which are motion blur (kernel size: $5$, angle: [$-45^\circ$, $45^\circ$]), black/blue/purple blocks ([$1/4$, $1$] of raw image size) and random scale ($-20$ to $+20\%$ on x- and y-axis independently)).
We randomly choose $m$ segments in a normal video sequence to augment and the augmented data are still treated normal.

Given the simulating noise data $\{\ddot\mcX^{(i)}\}|_{i=1}^{\ddot n}$ and the corresponding label set $\{\ddot y^{(i)}\}|_{i=1}^{\ddot n}$, we apply a supervised constraint on the predicted anomaly states $\{\ddot s_t^{(i)}\}|_{i=1}^{\ddot n}$:

\begin{align}
\zeta^{nse}&(\{{\ddot\mcX}^{(i)}\}|_{i=1}^{\ddot n}) = \frac{1}{\ddot n}\sum_{i=1}^{\ddot n}\sum_{t=1}^m ({\ddot s}_t^{(i)}-{\ddot y}^{(i)})^2,\\
&\qquad\st, ~{\ddot s}_t^{(i)} = \frac{1}{2}([{\ddot s}_t^{(i)}]^{sem} + [{\ddot s}_t^{(i)}]^{var}).
\end{align}

\textbf{Hand-crafted Anomaly}. The noise simulation strategy introduced above is able to alleviate the false alarm for normal videos. However, for anomaly videos, there still lacks enough data for model training. In particular, there exists no explicit location supervision in those anomalous videos which brings in great challenge for effective anomaly localization. 
To mitigate this issue, we then introduce hand-crafted anomaly to boost the anomaly localization performance via creating explicit location instructions for anomaly localization.
We name hand-crafted anomalies as pseudo-location data $\{\breve\mcX^{(i)}\}|_{i=1}^{\breve n}$.
Specifically, we first randomly choose a pair of normal and abnormal videos. Then several random segments of the normal video are selected and fused with several segments from the abnormal video. Finally, the obtained segments are combined with the remaining normal video segments to form a pseudo anomalous sequence. Those fused segments are targeted to be abrupt in the obtained sequence to create a pseudo anomalous sample since the substitutes differ from the distribution of the original normal video due to different scenes.

To decrease the abrupt in the fused sequence, we fused the features of abnormal segments with the normal ones. The coefficient of the anomaly feature ranges in [0.2,0.5] and were randomly generated during the training process. Rather than simply assigning a fixed score (e.g., $1$) for the simulated abnormal video will bring in a degenerate solution because the signal can encourage the remaining normal segments to have a high anomaly score along with pseudo-location data.
To mitigate the issue above, we propose a simple yet effective skill by barely pushing the fused segments $\{\breve\mcX_i\}|_{i=1}^{\breve n}$ to have a higher score than the others.
The supervision constraint is derived as:
\begin{align}
\zeta^{loc}(\{\breve\mcX_i\}|_{i=1}^{\breve n}) &= \frac{1}{\breve n}\sum_{i=1}^{\breve n}\sum_{t\in\mcI}~\max(0, {\breve s}_t^{(i)} - \max \limits_{j \notin \mcI}~\{{\breve s}_j^{(i)}\}),
\end{align}
where ${\breve s}_t^{(i)}$ denotes the anomaly score estimated by HCE module and $\mcI$ is a collection of the indexes for those pseudo location segments of the hand-crafted anomalies.
Integrating the above two augmentation techniques, the objective function of weak supervision enhancement strategy becomes:
\begin{align}
\label{eqn:zeta_a}
\zeta^{\mcA} =\zeta^{nse}(\{{\ddot\mcX}^{(i)}\}|_{i=1}^{\ddot n})+ \zeta^{loc}(\{\breve\mcX_i\}|_{i=1}^{\breve n}).
\end{align}

Combining Eqn.~\ref{eqn:zeta_o} and Eqn.~\ref{eqn:zeta_a}, we finally arrive at the overall objective function which is denoted by Eqn.~\ref{final_formula}.
\begin{figure*}[!t] 
	\centering 
	\includegraphics[width=1.0\textwidth]{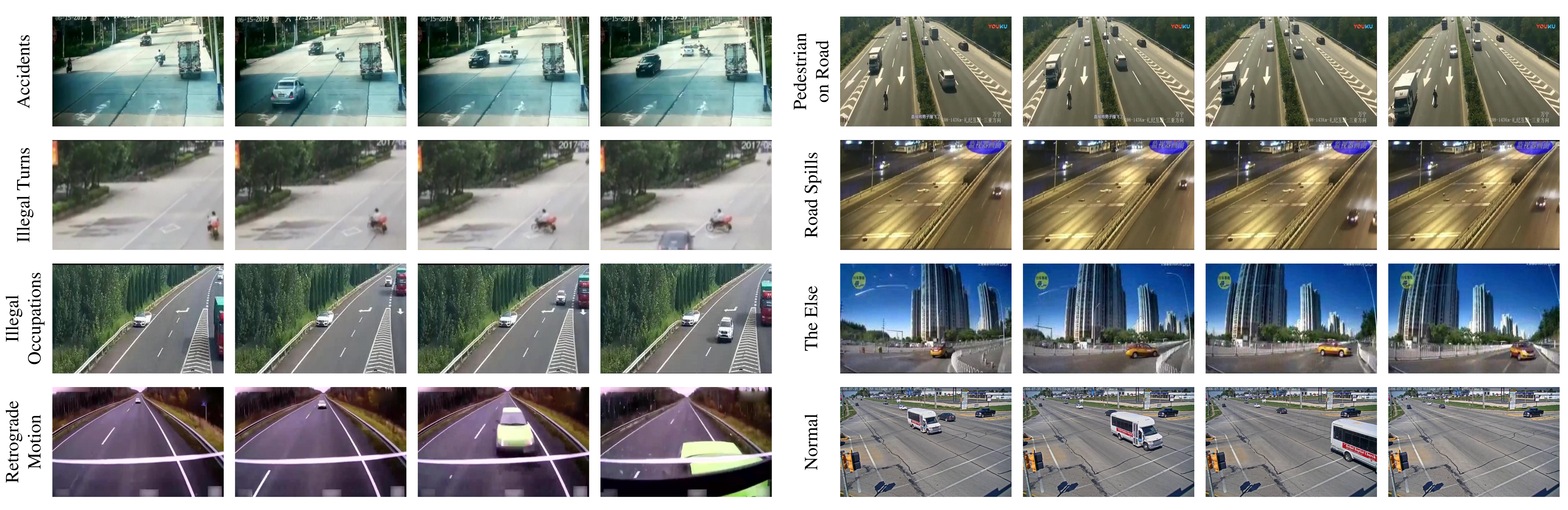}
	\caption{Examples of different anomalies in the collected TAD dataset. Best viewed in color.} 
	\label{Fig.examples} 
	\vspace{-0.3cm}
\end{figure*}

\subsection{Traffic Anomaly Detection (TAD) Dataset}
So far, most existing video anomaly datasets are prepared for unsupervised case, e.g., UCSD Pedestrian 1\&2 \cite{li2013anomaly}, Subway Entance \& Exit \cite{adam2008robust}, Avenue \cite{lu2013abnormal}, etc.
These unsupervised datasets are either small in scale or under the constraint of limited scenes.
For example, videos in Avenue are short and some of the anomalies are performed by actors (e.g., throwing paper), which are unrealistic.
Different from them, UCF-Crime \cite{sultani2018real} dataset is a newly released large-scale dataset proposed for weak supervision case.
Long untrimmed surveillance videos, covering 13 real-world anomalies, are collected in the dataset.
It has a total of $1,900$ surveillance videos, which consists of $1,610$ training videos and $290$ test videos.
Note that only video-level annotations are provided in the training set, and frame-level annotations are available for evaluation on the test set. The comparison of video anomaly detection datasets are shown in Table~\ref{datasets}.

Although the datasets mentioned above have greatly promoted the development of anomaly detection methods, there still lacks benchmarks of enough diversity for evaluation. To further meet the benchmark diversity requirement, we here propose a new anomaly detection dataset which specifies in the traffic scenes, differing greatly from the datasets mentioned before.
Traffic video monitoring plays an essential role in early warning and emergency assistance for car accidents.
It is an urgent need to design effective anomaly detection systems for surveillance videos on roads.
In traffic scenes, many factors, such as the vehicles moving at a high speed and various road conditions, add up to the hardness of anomaly detection.
So far, there is not any specific dataset for traffic anomaly detection. 
Although UCF-Crime contains road accidents videos, most of anomalies in traffic scenarios are not covered in this dataset.
Basically, a large-scale and complex dataset is of great importance for devising and evaluating various methods.
It is out desire to push the study of anomaly detection towards the usage in real traffic application.

To date, there are also public-available anomaly datasets on traffic scenes, however they are with limited scenarios. For example, in \cite{yao2018unsupervised}, a video dataset of unsupervised traffic accident detection is released and authors in \cite{chan2016anticipating} propose a fully supervised traffic anomaly detection benchmark. Both the above datasets are made up of samples of short clips, which can not represent the real situations that the anomalies rarely exist among a large quantity of normal data, and these datasets are not target for weak supervision. Moreover, these datasets only consist of first-person or dashboard videos, which lacks the videos captured by surveillance cameras on roads

Hence, we are motivated to construct a new large-scale dataset under the traffic scenes for video anomaly detection under weak supervision.
The collected TAD dataset consists of long untrimmed videos which cover $7$ real-world anomalies on roads, including \textit{Vehicle Accidents, Illegal Turns, Illegal Occupations, Retrograde Motion, Pedestrian on Road, Road Spills} and \textit{The Else} (i.e., the remaining anomalies with fewer quantity are put together as one category).
Some cases of the anomalies are shown in Figure~\ref{Fig.examples}.
The proposed dataset is comprehensive that includes realistic videos from various scenarios, weather conditions and daytime periods. 
\textcolor{blue}{
\begin{table}[!thbp]
	\centering
	\caption{A comparison of anomaly detection datasets}
	\scalebox{0.65}{
		\begin{tabular}{cccccc}
			\toprule
			Dataset &Target Domain& \# Videos &Total Frames & View & Supervision\\
			\midrule
			UCSD Ped 1/2 \cite{li2013anomaly}&Campus&98&18,560&3rd-person& Unsupervised\\
			\midrule
			CUHK Avenue \cite{lu2013abnormal}&Campus&37&30,652&3rd-person& Unsupervised\\
			\midrule
			Street scene \cite{ramachandra2020street}&Street&81&203,251&3rd-person& Unsupervised\\
			\midrule
			Shanghai Tech \cite{luo2017revisit}&Campus&81&317,398&3rd-person& Unsupervised\\
			\midrule
			UCF-Crime&General&1900&13,769,300&3rd-person&Weakly-supervised\\
			\midrule
			Aadv \cite{chan2016anticipating}&Traffic&1750&175,000&1st-person&Fully-supervised\\
			\midrule
			A3D \cite{yao2018unsupervised}&Traffic&1500&208,166&1st-person& Unsupervised\\
			\midrule
			\textbf{Ours} &\textbf{Traffic}&\textbf{500}&\textbf{540,272}&\textbf{1st\&3rd-person}&\textbf{Weakly-supervised}\\
			\bottomrule
	\end{tabular}}
	\label{datasets}
	\vspace{-0.3cm}
\end{table}
}

\textbf{Data collection.}
Traffic videos from various countries are collected and annotated under a detailed and unified plan.
Raw videos are downloaded from YouTube or Google website.
The collected videos are mostly recorded by CCTV cameras mounted on the roads.
We remove videos which fall into any of the following cases: manually edited, prank videos, and containing compilation.
Videos with ambiguous anomalies are also excluded.

\textbf{Data partition and Annotation.}
Our TAD dataset contains a total of about 25 hours videos, average 1075 frames per clips. The anomalies randomly occur in each clip, about 80 frames average and there are one to two random anomalies in a video sequence.
Finally, $500$ traffic surveillance videos are saved and annotated for anomaly detection, with 250 abnormal and normal videos respectively.
The whole dataset is randomly partitioned into two parts:
training set with $400$ videos, and test set with $100$ videos.
Both training and test sets contain normal and abnormal videos and all seven kinds of anomalies at various temporal locations in anomalous videos.
Following the setting of weak supervision as \cite{sultani2018real}, the training set is equipped with video-level annotations, and frame-level annotations are provided for the inference set.

Our proposed TAD dataset contains totally different abnormal scenarios than the current benchmarks. We believe that it could be used to better evaluate the effects of different anomaly detection algorithms from another perspective. We hope our TAD dataset could serve as a standard benchmark for better promoting the development of anomaly detection methods.

\section{Experiments}
\label{sec:exp}
\subsection{Implementation Details}
As in~\cite{zhong2019graph}, we adopt the Temporal Segment Network (TSN)~\cite{wang2018temporal}, which is a powerful action feature extractor, as our backbone net. We use the BN-Inception version of TSN to extract features for our proposed WSAL method.
We extract features from the global average pooling layer (1024-dim). For the UCF-Crime dataset, we use the model weights finetuned on UCF-Crime as in \cite{zhong2019graph} to extract features. While on our TAD dataset we only use the model weights pretrained on Kinetics-400 dataset. 
We first divide each video into $32$ non-overlapping segments empirically as in previous works~\cite{zhong2019graph,zhu2019motion,sultani2018real} for a fair comparison.
Hence, for each video, we have a $32$ $\times$ $1024$ feature matrix.
In our VAD model, the input features are first run through 2 fully connect layers with 512 and 128 dimension, respectively. Then, the dimensions of fully connect layers in HCE module and the immediate semantic score are 128 and 1, respectively. Dropout operations of 60\% rate are implemented after each fully connect layer, except the fully connect layer for immediate semantic score.
During the training phase, we randomly select 30 positive and 30 negative bags as a mini-batch.
We employ Adagrad \cite{duchi2011adaptive} optimizer with the initial learning rate of $0.001$.
The parameter of sparsity constraint in the margin loss is set to $\beta= 0.00008$ as in \cite{sultani2018real,zhu2019motion} and the weight of strategy for weak supervision enhancement is set to $\lambda = 1.0$ for the best performance.
We train the model for a total of $3$K iterations, decrease the learning rate by half at $1.2$K, $2.4$K and stop at $3$K. All hyper-parameters are the same for both UCF-Crime and TAD datasets.
\vspace{-0.3cm}
\subsection{Evaluation Metrics}
\label{EM}
For anomaly detection \cite{mahadevan2010anomaly,liu2018future}, Receiver Operation Characteristic (ROC) is used as a standard evaluation metric.
It is calculated by gradually changing the threshold of regular scores on the predicted anomaly scores.
Then the Area Under Curve (AUC) is accumulated to a score for the performance evaluation.
A higher value indicates a better anomaly detection performance.
Following the previous works~\cite{zhu2019motion,sultani2018real,zhong2019graph}, we apply ROC curves and frame-level AUC for anomaly detection performance comparison.
Due to the lack of frame-level annotations on the training split and verification split for ablation studies, we use video-level AUC as the measurement for tuning the hyper-parameters.
In addition, we also use the ROC and AUC on the anomaly subset to serve as the evaluation metric for anomaly localization ability.
\begin{figure*}[!t]
	\centering
	\includegraphics[width=0.9\textwidth]{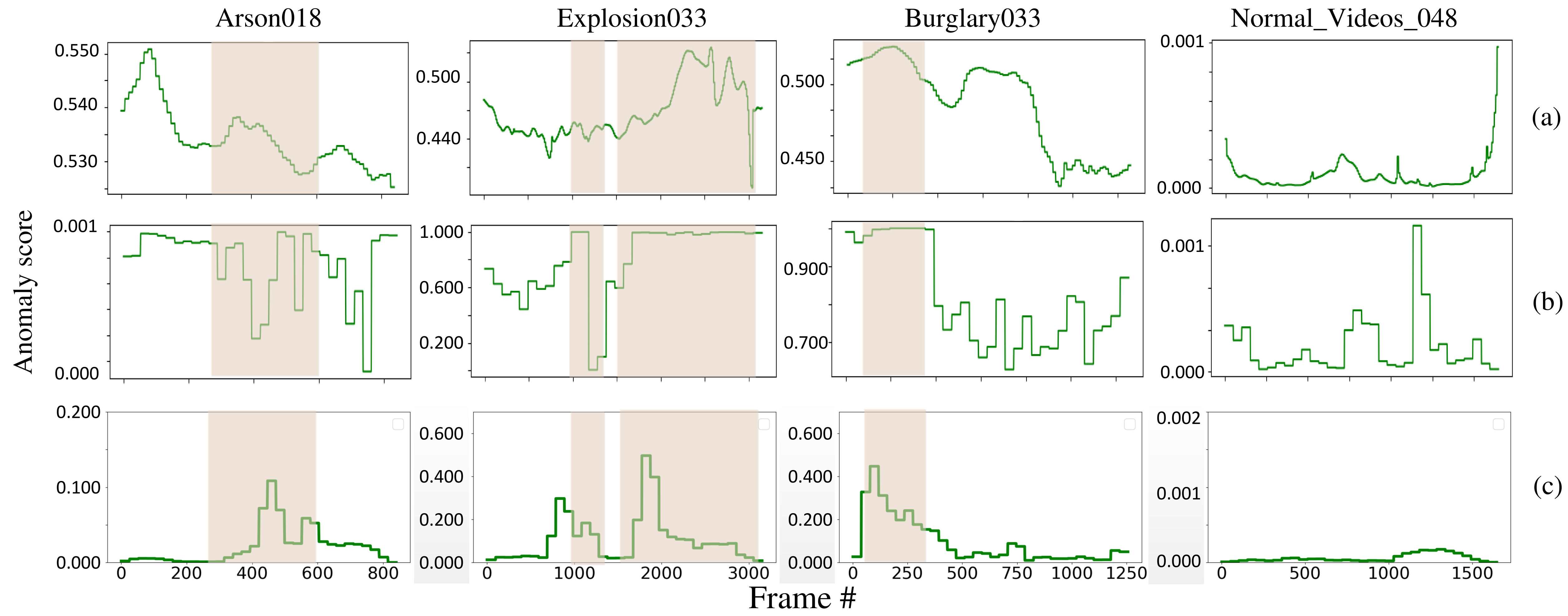}
	\caption{Visualization of predictions on the UCF-Crime test cases. The x-axis denotes the video frame \# and y-axis is corresponding to the anomaly score. In the figure, (a), (b), and (c) denotes the results of~\cite{sultani2018real}, \cite{zhong2019graph} and our model, respectively. The green curves are predictions of various approaches. The light orange regions are ground truth anomalies. Video names are labeled in the blank.}
	\vspace{-0.3cm}
	\label{testcases}
\end{figure*} 
\begin{figure}[t] 
    \label{ROCs} 
	\centering 
	\subfigure[ROC curves on UCF-Crime]{
		\centering
		\label{Fig.roc.sub.1}
		\includegraphics[width=0.23\textwidth]{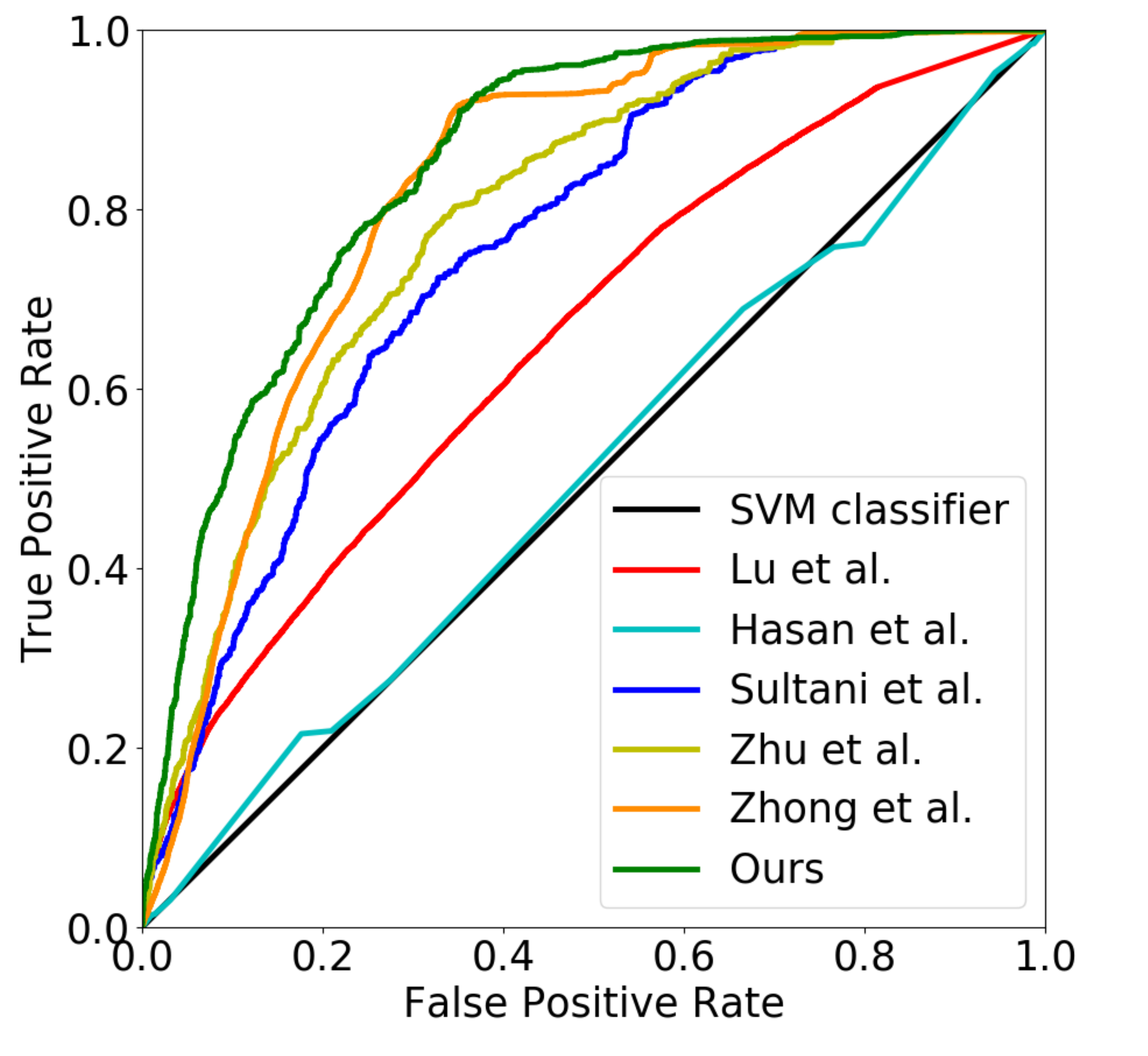}}
	\subfigure[ROC curves on our TAD]{
		\centering
		\label{Fig.roc.sub.2}
		\includegraphics[width=0.23\textwidth]{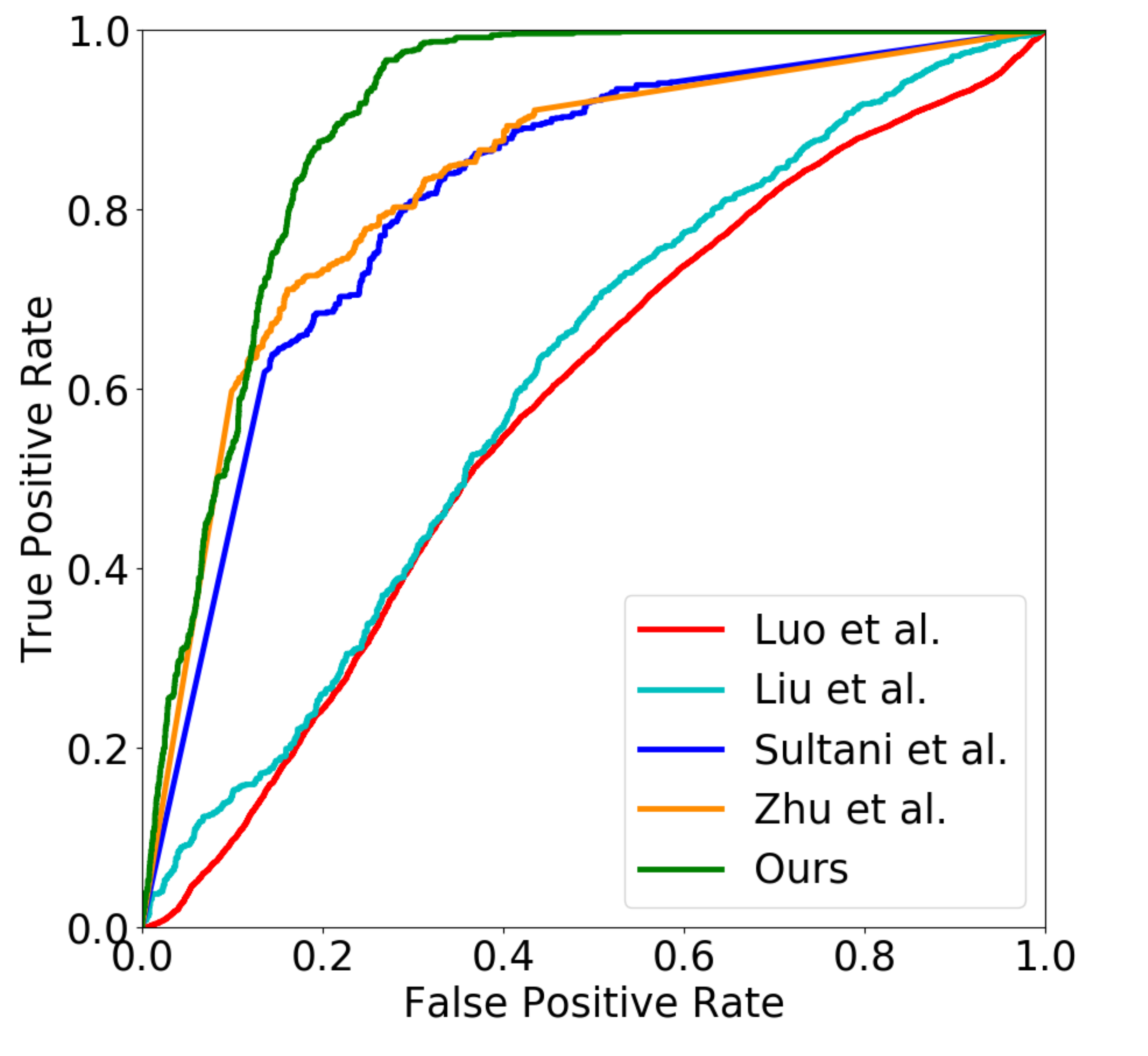}}
	\caption{ROC curves with various anomaly detection methods on the UCF-Crime dataset and TAD dataset}
    \vspace{-0.6cm}
\end{figure}
\subsection{Comparison with SOTA Methods}
\textbf{On UCF-Crime dataset.}
For fair comparison, we reproduce the methods of \cite{sultani2018real} and \cite{zhong2019graph} by running their publicly released codes. 
Other statistical results are drawn from the work \cite{sultani2018real}.
We compare our our WSAL with several anomaly detection methods.
Specifically, a binary SVM classifier is set as the baseline method. In this case, the anomalous and normal videos are treated as two separate class.
Models from Lu et al. \cite{lu2013abnormal} and Hasan et al. \cite{hasan2016learning} are two unsupervised methods, training with the normal videos in UCF-Crime training set.
The remaining Sultani et al. \cite{sultani2018real}, Zhu et al. \cite{zhu2019motion} and Zhong et al. \cite{zhong2019graph} are SOTA weakly-supervised methods.
As shown in Table~\ref{SOTA}, on the whole test set which contains both the normal and abnormal videos, we boost the best performance of overall AUC from the $82.12$\% to $85.38$\% by a large margin. 
In Figure \ref{Fig.roc.sub.1}, we plot the ROC curves of SOTA methods on the whole UCF-Crime dataset and it vividly shows the superoirity of our proposed WSAL method over other SOTA methods.
As for the Anomaly subset, our proposed method exceeds the SOTA detectors by $9$\% over \cite{zhong2019graph} and $13$\% over \cite{sultani2018real}, achieving a significant progress on the anomaly localization perspective. 

We draw the following conclusions upon above experimental results: 1) SVM classifier fails to distinguish the anomalous and normal videos, mainly because the normal patterns take the dominate position in both normal and anomalous videos, and make the classifier difficult to capture the rare anomalies; 2) By encoding the normal patterns and building the corresponding semantic boundary, unsupervised methods \cite{hasan2016learning} and \cite{lu2013abnormal} achieve better results than SVM classifier; 3) Owing to the benefits of weak labels, the performances of weakly-supervised methods \cite{sultani2018real}, \cite{zhu2019motion} and \cite{zhong2019graph} are superior than above approaches. Nevertheless, previous weakly-supervised methods infer the anomaly status from high-level semantic features intuitively, while neglecting an important property of the anomaly, which is the dynamic evolution lying in time series. 
The considerable gain in anomaly localization promotes the improvement of overall anomaly detection accuracy.
Some visual results of our method on test cases are shown in Figure \ref{testcases}. Compared with the other two SOTA methods, a large degree of distinction between the normal and anomalous can be achieved by our approach. As a result, a superior performance is achieved with the better anomaly-discriminating capability.

\begin{table}[!tbp]
	\centering
	\caption{Quantitative comparison on the UCF-Crime dataset. * symbol indicates the method is trained with normal videos only}
	\scalebox{0.8}{
		\begin{tabular}{ccccc}
			\toprule
			Method &Overall AUC(\%)&Anomaly Subset AUC(\%)&&\\
			\midrule
			SVM &50&50\\
			\midrule
			Hasan et al.* \cite{hasan2016learning}&50.60&-\\
			Lu et al.* \cite{lu2013abnormal}&65.51&-\\
			\midrule
			Sultani et al. \cite{sultani2018real}&75.41&54.25\\
			Zhu et al. \cite{zhu2019motion}&79.10&62.18\\
			Zhong et al. \cite{zhong2019graph} &82.12&59.02\\
			\textbf{Ours} &\textbf{85.38}&\textbf{67.38}\\
			\bottomrule
	\end{tabular}}	
	\label{SOTA}
\end{table}

\textbf{On the proposed TAD dataset.}
To compare the performance of different methods under other circumstances, we conduct comparison experiments on the TAD dataset.
We compare our WSAL model with four SOTA anomaly detection methods, including two unsupervised methods (Luo et al. \cite{luo2017revisit} and Liu et al. \cite{liu2018future}) and two weakly-supervised methods (Sultani et al. \cite{sultani2018real} and Zhu et al. \cite{zhu2019motion})
~For unsupervised models, we follow their implementation and train the models on the training subset where only normal videos are provided.
All models are re-trained with the same features extracted using TSN, except \cite{liu2018future} which takes the RGB frames as inputs.

The quantitative comparisons of AUC are revealed in Table~\ref{traffic} and the corresponding ROC curves are drawn in Figure~\ref{Fig.roc.sub.2}.
Similar as upon UCF-Crime, weakly-supervised methods are able to obtain much better performance than unsupervised ones. 
As both normal and abnormal training samples are provided, the weakly-supervised methods own much better understanding of the intrisic nature of anomaly. 
It demonstrates that 1) weak annotations with a little cost of time and work can greatly benefit the performance of VAD methods, and 2) the collected dataset with various scenes and different kinds of anomalies are comlex and extremely challenging for unsupervised methods, compared with current unsupervised benchmark, such as Ped 1/2, Avenue, etc., which could activate future research direction for unsupervised methods.
In addition, our WSAL method also achieves better performance with a gain of $6$\% AUC over previous SOTA~\cite{zhu2019motion}. 
The prominent advances on the two large-scale and comprehensive benchmarks prove the superiority of our method on detecting and localizing anomalies.
\begin{figure}[t]
	\centering
	\includegraphics[width=0.4\textwidth]{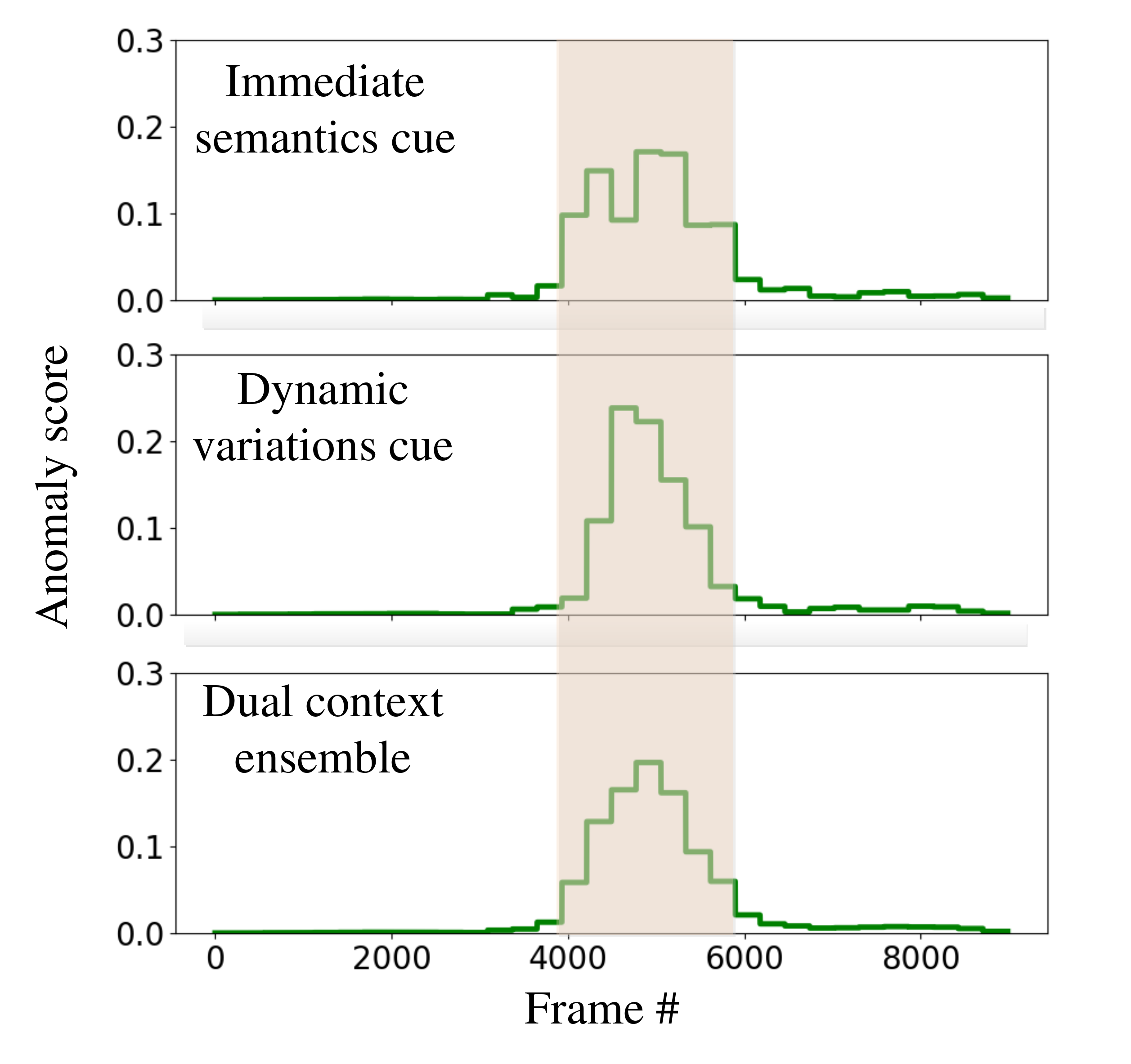}
	\caption{An visualization case of the dual context cues on the UCF-Crime dataset. The light orange region denotes the groundtruth anomaly. From the top to the bottom, the curves represent the anomaly scores of immediate semantics cue, dynamic variations cue and consensus of above two cues, respectively. A more robust and smooth prediction is observed from the dual context model.}
	\label{Fig.DCE}
	\vspace{-0.3cm}
\end{figure}
\vspace{-0.3cm}
\subsection{Ablation Studies}
To comprehensively study the impact of different components we proposed, we conduct various ablation studies in this part. All experiments are conducted on the UCF-Crime dataset, all hyper-parameters are kept the same as the WSAL method if not otherwise claimed.

\textbf{Analysis of the dual context ensemble.}
To verify the effectiveness of the proposed ensemble mechanism of immediate semantics and dynamic variations, we construct three variants of the proposed WSAL method where only the immediate semantics or the dynamic variations or both of them are adopted. Detailed comparisons are presented in the first three lines of Table~\ref{Ablations}. When only immediate semantics are exploited, the algorithm is able to achieve satisfactory accuracies of $81.44$\% and $61.13$\% w.r.t anomaly detection and anomaly localization. It means that the immediate semantics can provide useful information for the tasks, and if dynamic variations are adopted, the performances are boosted by $1.12$\% and $1.25$\% separately. One case for visualization is shown in Figure~\ref{Fig.DCE}.
There is only one clear and sharp peak in the prediction of dynamic variations cue, compared with the results of immediate semantics cue.
It demonstrates that the dynamic variations are good at capturing the sudden occurrence of the anomaly even when the immediate semantics cue may bring in uncertainty. By aggregating the immediate semantics and dynamic variations cues, the detection performance is more robust under various circumstances, with a higher detection and localization accuracy. Thanks to the complementary characteristic of the immediate semantics and dynamic variations cue, which stand for different aspects of the anomaly.

\begin{table}[t]
	\centering
	\caption{Quantitative comparison on our TAD dataset. * symbol indicates the method is trained with normal videos only}
	\scalebox{1.0}{
		\begin{tabular}{ccc}
			\toprule
			Method&Overall AUC(\%)&Anomaly Subset AUC(\%)\\
			\midrule
			Luo et al.* \cite{luo2017revisit}&57.89&55.84\\
			Liu et al.* \cite{liu2018future}&69.13&55.38\\
			Sultani et al. \cite{sultani2018real}&81.42&55.97\\
			Zhu et al. \cite{zhu2019motion}&83.08&56.89\\
			\midrule
			$\text{Ours}$&\textbf{89.64}&\textbf{61.66}\\
			\bottomrule
	\end{tabular}}
	\label{traffic}
	\vspace{-0.3cm}
\end{table}
\begin{table}[t]
	\centering
	\caption{Ablation studies of our WSAL method on the UCF-Crime dataset. The meanings of the abbreviations in the table are as follows: IS: Immediate Semantics; DV: Dynamic Variations; HCE: High-order Context Encoding; NS: Noise Suppression; HA: Hand-crafted Anomaly}
	\setlength{\tabcolsep}{3mm}{
		\scalebox{0.8}{
			\begin{tabular}{ccccccc}
				\toprule
				IS&DV&HCE&NS&HA&Overall AUC(\%)&Anomaly Subset AUC(\%)\\
				\midrule
				\checkmark& & & & &81.44&61.13\\
				&\checkmark & & & &82.52&62.38\\
				\checkmark&\checkmark& & & &82.95&63.65\\
				\checkmark&\checkmark&\checkmark& & &84.44&64.95\\
				\checkmark&\checkmark&\checkmark&\checkmark& &84.86&66.28\\
				\checkmark&\checkmark&\checkmark& &\checkmark&84.95&66.55\\
				\midrule
				\textbf{\checkmark}&\textbf{\checkmark}&\textbf{\checkmark}&\textbf{\checkmark}&\textbf{\checkmark}&\textbf{85.38}&\textbf{67.38}\\
				\bottomrule
	\end{tabular}}}
	\vspace{-0.3cm}
	\label{Ablations}
\end{table}

\textbf{Analysis of HCE model.} 
We study the influence of High-order Context Encoding on the new training and verification splits of UCF-Crime. Since only video-level labels are available in verification split, video-level AUC is measured by aggregating the segment-level model predictions as in Formula \ref{f1} and then calculating the AUC results.
As to the incorporation of temporal context, an appropriate temporal window size $k$ is critical for the final performance. We slowly increase the window size $k$ from 0 to 3 and the results are listed in Table~\ref{HDR}. When the window size $k$ grows, the accuracy of video-level predictions improves drastically from 0 to 1, with a performance gain of 1.6\%.
It means that appropriate aggregation of the temporal context possesses great potential for anomaly detection. The fruitful information in the temporal neighborhood facilitates the learning of anomaly semantics, as well as the encoding of the temporal evolution.
Finally, we choose the size $k=2$ for trading off between model size and performance, since the accuracy gain slows down when the window size further increases.

Further, We conduct ablation studies for analysing impacts of high-order context encoding on various feature sources as in Table.~\ref{feature}. For feature extraction on I3D~\cite{carreira2017quo}, we implement official released model; On R(2+1)D~\cite{Tran_2018_CVPR}, we use the official 18 layer model, both the models are pre-trained on kinetics-400. As is shown, when the high-order context encoding module is implemented with input feature of I3D and R(2+1)D, the anomaly location accuracy is increased by over 3\% (from 51.23\% to 55.04\% with I3D inputs and from 47.96\% to 51.25\% with R(2+1)D inputs), together with about 2\% on overall accuracy. Moreover, based on the TSN feature pre-trained on UCF-Crime (by tiling video-level label for each video segment as in \cite{zhong2019graph}), the proposed HCE module can still achieve remarkable progress with overall AUC from 82.95\% to 84.44\%. It demonstrates that although temporal aggregation has been done in the feature extraction model, our high-order context encoding is still beneficial for highlighting and capturing transient anomalous information in long-time sequences. It boosts the performance of video anomaly detection.

\begin{figure}[t]
	\centering
	\includegraphics[width=0.5\textwidth]{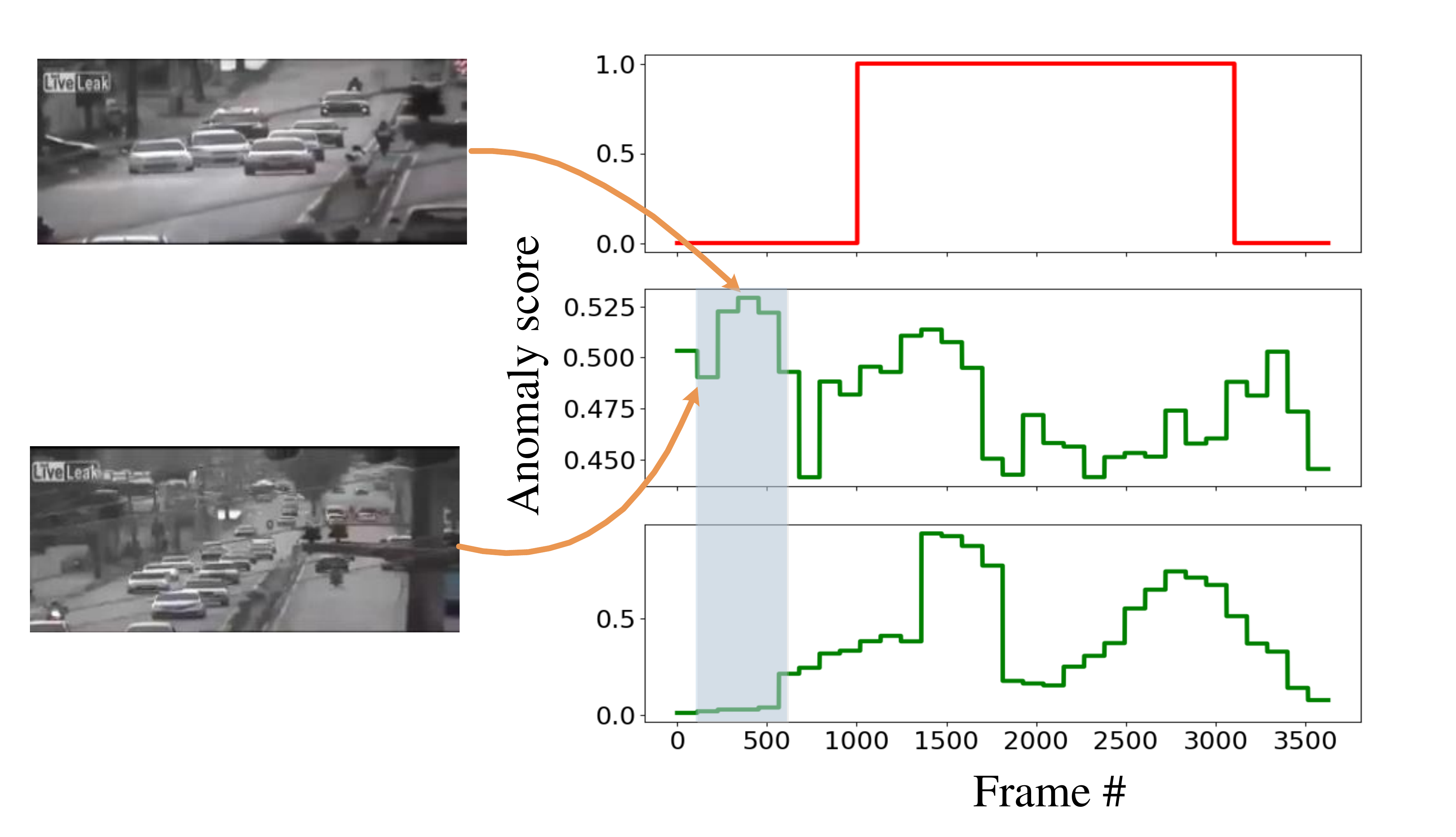}
	\caption{An visualization case of video noises on the UCF-Crime dataset. The light blue region in the video sequence contains video noises. The red curves in the top-right denotes the groundtruth anomalies. The curve in the middle-right represents the result of baseline method \cite{zhong2019graph}. The bottom-right curve belongs to the result of our method. In this case, the noise comes from the lens jitters. The drastic view change easily leads to the false detection of the basic model.} 
	\label{Fig.noise} 
\end{figure}

\begin{table}[t]
	\centering
	\caption{Analysis of the window size in HCE model on the UCF-Crime dataset}
	\setlength{\tabcolsep}{3mm}{
		\scalebox{1.0}{
			\begin{tabular}{ccccc}
				\toprule
				Window Size &0&1&2&3\\
				\midrule
				Video-level AUC(\%)&93.39&95.01&95.65&95.73\\
				\bottomrule
	\end{tabular}}}
	\label{HDR}
	\vspace{-0.5cm}
\end{table}
\textcolor{blue}{
\begin{table}[t]
	\centering
	\caption{Ablation studies of various feature sources on UCF-Crim e benchmark. + symbol indicates the base model is equipped with high-order context encoding module.}
	\scalebox{1.0}{
		\begin{tabular}{ccc}
			\toprule
			Feature Source&Overall AUC(\%)&Anomaly Subset AUC(\%)\\
			\midrule
			I3D~\cite{carreira2017quo}&72.43&51.23\\
			I3D + &74.18&55.04\\
			R(2+1)D~\cite{Tran_2018_CVPR} &72.47&47.96\\
			R(2+1)D + &75.59&51.25\\
			\midrule
			TSN~\cite{wang2018temporal} &82.95&63.65\\
			TSN + &84.44&64.95\\
			\bottomrule
	\end{tabular}}
	\label{feature}
	\vspace{-0.4cm}
\end{table}
}
\vspace{-0.1cm}
\textbf{Enhanced Weak Supervision}
We conduct studies to verify the effectiveness of the proposed video noise augmentation and hand-crafted anomaly separately. The detailed results are reported in Table \ref{Ablations}. We first augment the training process by adding video noise simulations. The anomaly detection AUC is boosted from $84.44$\% to $84.86$\% and the anomaly localization accuracy is further boosted by $1.33$\%. One case is plotted in Figure \ref{Fig.noise}, our noise stimulation strategy contributes to alleviate the interference caused by lens jitters. Note that the anomaly localization performance improvement is non-trivial. It clearly demonstrates that the proposed noise augmentation strategy is able to aid the dynamic variation module for better capturing the real anomaly and achieving better understanding of the intrinsics of anomalies.
When we manually synthetic some anomalies to aid the training process, our method achieves $0.51$\% and $1.60$\% performance gain over the training strategy where no noise augmentations are adopted. The performance promotions demonstrate that our synthetic anomaly data is able to provide extra useful supervision, indicating that larger abnormal detection dataset is needed for sufficient training of abnormal detection methods. If both augmentation strategies are combined, the proposed method is able to achieve much better performance than the two separate augmentation strategies. It indicates that the proposed two augmentation strategies are beneficial for the understanding of the anomaly concept by suppressing the interference coming from the environment as well as hardware failures, and generating pseudo signals that simulating the occurrence of anomalies. 

\textbf{Speed Analysis}
The whole model can run at 44 FPS on a single RTX 2080Ti GPU.
Among them, the feature extraction model----TSN with BN-Inception backbone~\cite{wang2018temporal} runs at 45 FPS, with the input frame resolution set as 224x224. 
Then, based on the extracted feature, it only consumes 1.81 ms (runs at 550 FPS) to predict the anomaly scores.
In summary, our model is applicable in the on-line applications.

\section{Conclusion}
\label{sec:con}
In this work, we focused on anomaly localization in surveillance videos and proposed a weakly supervised anomaly localization network that deeply exploring the temporal context in consecutive segments.
Our model encoded temporal dynamic variations as well as high-level semantic information, and leveraged both of them for anomaly detection and localization.
Furthermore, we devised a weak supervision enhancement strategy.
The accuracy of anomaly localization was greatly improved under the introduced supervision of video noise augmentation and pseudo-location data.
We also collected a new traffic anomaly detection dataset for evaluating methods under realistic scenarios on roads. 
SOTA methods were verified on UCF-Crime dataset and our TAD dataset. The experimental results showed that the proposed anomaly detector has performed significantly better than previous methods.
\vspace{0.2cm}

\textbf{Acknowledgments.} This work was supported by the Natural Science Foundation of Jiangsu Province (Grant No. BK20190019), the National Natural Science Foundation of China (Grants Nos. 62072244, 61972204), and the Natural Science Foundation of Shandong Province (Grant No. ZR2020LZH008). This work was partly collaborated with State Key Laboratory of High-end Server \& Storage Technology.
{
\bibliographystyle{IEEEtran}
\bibliography{sample-base}
\footnotesize

\end{document}